\documentclass[10pt,twocolumn,letterpaper]{article}

\usepackage[pagenumbers]{cvpr} 

\usepackage[dvipsnames]{xcolor}

\usepackage{booktabs}       
\usepackage{amsfonts}       
\usepackage{nicefrac}       
\usepackage{microtype}      
\usepackage{amsmath}
\usepackage{graphicx,psfrag,epsf}
\usepackage{enumerate}
\usepackage{amssymb,amsfonts,mathtools}

\usepackage{algorithm}
\usepackage[tableposition=top]{caption}
\usepackage{algorithmic}
\usepackage{multirow}
\usepackage{cuted}
\usepackage{verbatim}
\usepackage{wrapfig}
\usepackage{varwidth}
\usepackage{parskip}
\usepackage{amsthm}

\makeatletter
\def\th@plain{%
	\thm@notefont{}
	\itshape 
}
\def\th@definition{%
	\thm@notefont{}
	\normalfont 
}
\makeatother

\makeatletter
\def\endthebibliography{%
	\def\@noitemerr{\@latex@warning{Empty `thebibliography' environment}}%
	\endlist
}
\makeatother

\usepackage{footnote}
\usepackage{wrapfig}
\usepackage{nicefrac}

\makesavenoteenv{tabular}

\newcommand{\rom}[1]{\uppercase\expandafter{\romannumeral #1\relax}}
\DeclarePairedDelimiterX{\norm}[1]{\lVert}{\rVert}{#1}
\DeclarePairedDelimiterX{\bnorm}[1]{\biggl\lVert}{\biggr\rVert}{#1}
\DeclarePairedDelimiterX{\abs}[1]{\lvert}{\rvert}{#1}

\renewcommand{\emph}[1]{{\textit{#1}}}

\newtheorem{definition}{Definition}
\theoremstyle{definition}
\newtheorem{remark}{Remark}
\newtheorem{theorem}{Theorem}

\newtheorem{lemma}{Lemma} 

\def\D{\mathcal{D}} 

\def\de{\overset{\Delta}{=}}

\def\de{\overset{\Delta}{=}}

\usepackage{pifont}

\definecolor{cvprblue}{rgb}{0.21,0.49,0.74}
\usepackage[pagebackref,breaklinks,colorlinks,citecolor=cvprblue]{hyperref}

\def\paperID{15557} 
\def\confName{CVPR}
\def\confYear{2024}

\title{RAW: A Robust and Agile Plug-and-Play Watermark Framework for AI-Generated Images with Provable Guarantees}

\author{
  Xun Xian\\
  University of Minnesota\\
  {\tt\small xian0044@umn.edu}
  \and
  Ganghua Wang\\
  University of Minnesota\\
  {\tt\small wang9019@umn.edu}
  \\ 
  \and
  Xuan Bi\\
  University of Minnesota\\
  {\tt\small xbi@umn.edu}
  \and
  Jayanth Srinivasa\\
  Cisco Research\\
  {\tt\small jasriniv@cisco.com}
  \and
  Ashish Kundu\\
  Cisco Research\\
  {\tt\small ashkundu@cisco.com}
  \\ 
  \and
  Mingyi Hong\\
  University of Minnesota\\
  {\tt\small mhong@umn.edu}
  \and
  Jie Ding\\
  University of Minnesota\\
  {\tt\small dingj@umn.edu}
}

\begin{document}
\maketitle
\begin{abstract}
Safeguarding intellectual property and preventing potential misuse of AI-generated images are of paramount importance. This paper introduces a robust and agile plug-and-play watermark detection framework, referred to as RAW.
As a departure from traditional encoder-decoder methods, which incorporate fixed binary codes as watermarks within latent representations, our approach introduces learnable watermarks directly into the original image data. Subsequently, we employ a classifier that is jointly trained with the watermark to detect the presence of the watermark.
The proposed framework is compatible with various generative architectures and supports on-the-fly watermark injection after training. By incorporating state-of-the-art smoothing techniques, we show that the framework also provides provable guarantees regarding the false positive rate for misclassifying a watermarked image, even in the presence of certain adversarial attacks targeting watermark removal. Experiments on a diverse range of images generated by state-of-the-art diffusion models show substantial performance enhancements compared with existing approaches. For instance, our method demonstrates a notable increase in AUROC, from 0.48 to 0.82, when compared to state-of-the-art approaches in detecting watermarked images under adversarial attacks, while 
maintaining image quality, as indicated by closely aligned FID and CLIP scores. 
\end{abstract}    

\section{Introduction}

In recent years, Generative Artificial Intelligence has made substantial progress in various fields. Notably, in computer vision, the introduction of diffusion model (DM) based applications such as Stable Diffusion~\cite{rombach2022high} and DALLE-2~\cite{ramesh2022hierarchical} has significantly improved the quality of image generation. These models exhibit the capacity to generate a wide spectrum of creative visuals, spanning both artistic compositions and realistic depictions of real-world scenarios. However, these exciting new developments also raises concerns regarding potential misuse, particularly in the malicious creation of deceptive content, as exemplified by DeepFake~\cite{verdoliva2020media}, and instances of copyright infringement~\cite{sag2023copyright}, which can readily replicate unique creative works without proper authorization.

To mitigate the potential misuse of diffusion models, the incorporation of watermarks emerges as an effective solution. Watermarked images, subtly tagged with imperceptible signals, act as markers, revealing their machine-generated origin. This documentation not only assists platforms and organizations in addressing concerns but also facilitates collaboration with law enforcement in tracing image sources~\cite{bender2021dangers}.
Watermarking techniques designed for generative models can be generally classified into two primary categories: model-agnostic~\cite{zhang2019robust,cox1996secure} and model-specific methods~\cite{fernandez2023stable, kim2023wouaf, wen2023tree}.
Model-specific approaches are closely tied to specific generative models and frequently involve adjustments to various components of these models, which can possibly limit their flexibility and suitability for various use cases.
For instance, the Tree-Ring watermark~\cite{wen2023tree} is tailored for specific samplers, e.g., DDIM~\cite{song2020denoising}, used for image generation within diffusion models. The feasibility of adapting this method to other commonly used samplers remains unclear.

In contrast, model-agnostic approaches directly watermark generated content without modifying the generative models.
These approaches can be categorized into two types. The first, from traditional signal processing, e.g., DwTDcT~\cite{cox2007digital}, embeds watermarks in specific parts of images' frequency domains. However, they are vulnerable to strong image manipulations and adversarial attacks for removing watermarks~\cite{balle2018variational}.
The second type leverages deep learning techniques, using encoder-decoder structures to embed watermarks, e.g., binary codes, in latent spaces. For example, RivaGan~\cite{zhang2019robust} jointly trains the watermark and watermark decoder as learned models, enhancing transmission and robustness. Nonetheless, these methods require more computational resources for watermark injection, making them less suitable for real-time on-the-fly deployment.

Furthermore, due to possible economic consequences linked to the utilization of watermarks, such as unauthorized copying leading to subsequent financial losses, there has been a sustained emphasis on the importance of accurately measuring false-positive rates (FPR) and/or the Area Under the Receiver Operating Characteristic curve (AUROC) for every employed watermarking strategy~\cite{pitas1998method}.
To establish an explicit theoretical formulation for FPR, many studies have assumed that the binary watermark code extracted from unwatermarked images exhibits a pattern where each digit is an independent and identically distributed (IID) Bernoulli random variable with a parameter of 0.5. 
This assumption enables the explicit derivation of the FPR when comparing the extracted binary code with the predefined actual binary watermark code.
However, such an assumption may not hold as empirically observed in~\cite{fernandez2022watermarking}, and thus the corresponding formulation for FPR could be incorrect. Moreover, to our knowledge, none of these methods have provided \textit{provable} guarantees on FPR, even if the assumptions are met.

\subsection{Contributions}
In this paper, we introduce a \textbf{\underline{R}}obust, \textbf{\underline{A}}gile plug-and-play \textbf{\underline{W}}atermark framework, abbreviated as \textbf{RAW}.
RAW is designed for both adaptability and computation efficiency, providing a model-agnostic approach for real-time, on-the-fly deployment of image watermarking. This dedication to adaptability extends to ensuring accessibility for third-party users, encompassing artists and generative model providers. Moreover, this adaptability is fortified by the integration of state-of-the-art smoothing techniques for achieving provable guarantees on the FPR for detection, even under moderate adversarial attacks.

\begin{figure}
    \centering
    \includegraphics[width = 0.95
    \linewidth]{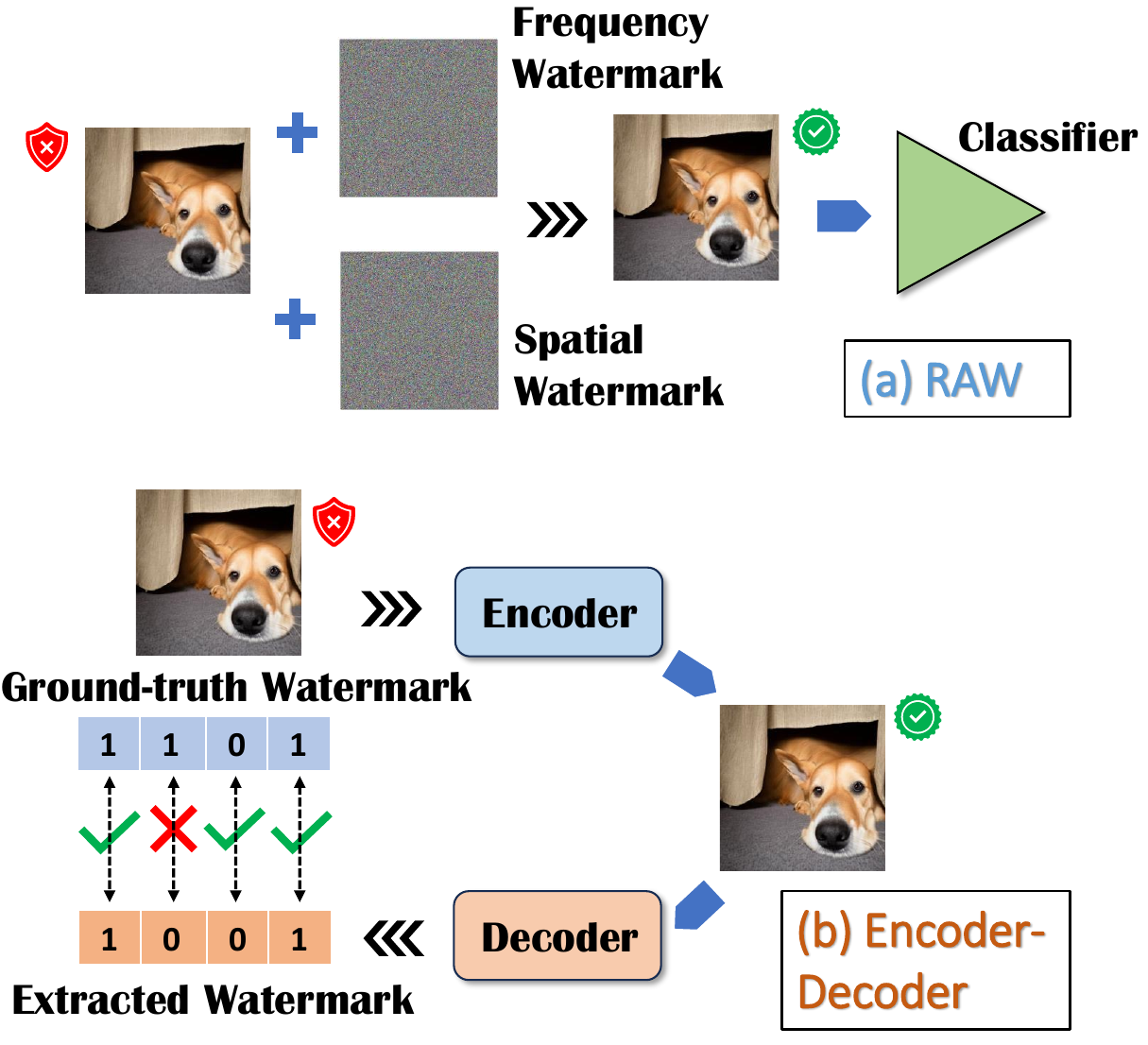}
    \caption{Illustration of our proposed RAW (top row) and popular encoder-decoder based watermarking schemes (bottom row).}
    \label{fig: high-level contrast}
    \vspace{-0.35cm}
\end{figure}

\textbf{A new framework for robust and agile plug-and-play watermark learning.}
As illustrated in Figure~\ref{fig: high-level contrast}, in contrast to encoder-decoder techniques that insert fixed binary watermarks into latent spaces, we propose to embed learnable watermarks, matching the image dimensions, into both the frequency and spatial domains of the original images. To differentiate between watermarked and unwatermarked samples, we utilize a classifier, e.g., a convolutional neural network (CNN), and perform joint training for both the watermarks and the classifier.
The proposed framework offers several benefits, including enhanced computational efficiency through batch processing for watermark injection post joint training, and it can be readily integrated/adapted with other state-of-the-art techniques to further enhance robustness and generalizability, such as adversarial training~\cite{goodfellow2014explaining, madry2017towards}, contrastive learning~\cite{chen2020simple, khosla2020supervised}, and label smoothing~\cite{muller2019does}.

\textbf{Provable guarantees on FPR even under adversarial attacks.}
By integrating advanced methods from the conformal prediction literature~\cite{vovk2005algorithmic,lei2014distribution} into our RAW framework, we showcase its ability to offer rigorous, distribution-free assurances regarding the FPR.
Additionally, we develop a novel technique, inspired by the randomized smoothing~\cite{duchi2012randomized, cohen2019certified}, to further enhance our provable guarantees. This extension ensures \textit{certified} guarantees on FPR under arbitrary perturbations with bounded norms, that is,  as long as any transformations or adversarial attacks on future test images stay within a predefined range, 
our FPR guarantees remain valid.


\textbf{Extensive empirical studies on various datasets.}
We evaluate the efficacy of our proposed method across various generative data scenarios, such as the DBDiffusion~\cite{wang2022diffusiondb} and the generated MS-COCO~\cite{lin2014microsoft}. Our assessment includes detection performance, robustness against image manipulations/attacks, the computational efficiency of watermark injection, and the quality of generated images. The experimental results consistently affirm the excellent performance of our approach, as evidenced by notable enhancements in AUROC from 0.48 to 0.82 under state-of-the-art diffusion-model-based adversarial attacks aimed at removing watermarks.

\subsection{Related Work}

\textbf{Classical watermarking techniques for images.}
Image watermarking has long been a fundamental problem in both signal processing and computer vision literature~\cite{cox1996secure,pereira1999template}. Methods for image-based watermarking typically operate within either the spatial or frequency domains~\cite{ cox1996secure, o1997rotation, altun2009optimal, gupta2016review}. Within the spatial domain, methodologies span from basic approaches, such as the manipulation of the least significant bit of individual pixels, to more complex strategies like Spread Spectrum Modulation~\cite{altun2009optimal, hartung1998watermarking} and Quantization Index Modulation~\cite{jie2009new}. In the frequency domain, watermark embedding~\cite{cox1996secure,o1997rotation} involves modifying coefficients generated by transformations such as the Discrete Cosine Transform~\cite{hernandez2000dct} and Discrete Wavelet Transform~\cite{kadu2016discrete, kothari2012transform}. These frequency transformations share the advantage of being able to handle basic image manipulations like translations, rotations, and resizing, while also enabling the construction of watermarks with resilience to these alterations. However, empirical evidence reveals their vulnerability to adversarial attacks and intense image perturbations, including rotations exceeding $90^\circ$~\cite{balle2018variational}.

\textbf{Image watermarking using deep learning.}
In recent times, the advent of advanced deep learning techniques has opened up new avenues for watermarking. Many of these methods~\cite{kandi2017exploring,hayes2017generating, zhu2018hidden, zhang2019robust, fernandez2022watermarking}, are based on the encoder-decoder architecture. In this model, the encoder embeds a binary code into images in latent representations, while the decoder takes an image as input and generates a binary code for comparison with the binary code injected for watermark verification.
For example, the HiDDeN technique~\cite{zhu2018hidden} involves the simultaneous training of encoder and decoder networks, incorporating noise layers specifically crafted to simulate image perturbations.
While these methods enhance robustness compared to traditional watermarking, they may not be ideal for real-time, on-the-fly watermark injection due to the time-consuming feed-forward process in the encoder, particularly with larger architectures.

\textbf{Watermarks for protecting model intellectual property}
Deep neural networks have emerged as valuable intellectual assets, given the substantial resources required for their training and data collection processes~\cite{rombach2022high}.
For instance, training the stable diffusion models requires roughly 150,000 GPU hours, at a cost of around \$600,000~\cite{StableDiffusionWiki}.
With their diverse applications in real-world scenarios, ensuring copyright protection and facilitating their identification is essential for both normal and adversarial contexts~\cite{tramer2016stealing,xian2022understanding}.
One approach aims to embed watermarks directly into the model parameters~\cite{uchida2017embedding, li2021spread} but necessitates white-box access to inspect these watermarks. In an alternative category of watermarking techniques~\cite{adi2018turning, zhang2018protecting, le2020adversarial, zhao2023recipe}, which rely on techniques called backdoor attacks~\cite{gu2017badnets, xian2023understanding}, backdoor triggers are injected into training data during the model training stage, e.g., an image of cat with a square patch positioned at the lower-right corner. During the testing phase, entities seeking ownership of the deep networks can present the backdoored inputs to the backdoored deep networks, enabling them to make targeted predictions on those inputs, e.g., consistently predicting cat images with a square patch positioned at the lower-right corner as a dog.  The central aim of watermarking for these works revolves around safeguarding the intellectual property of {models}, rather than protecting the generated outputs.

\section{Preliminary}

\textbf{Notations.}
We consider the problem of embedding watermarks into images and then detecting the watermarks as a binary classification problem.
Let $\mathcal{X} = [0,1]^{C \times W \times H}$ be the input space, with $C$, $W$, and $H$ being the channel, width and height of images, respectively. We denote $\mathcal{Y}=\{0,1\}$ to be the label space, with label 0 indicating unwatermarked and 1 indicating watermarked versions, respectively. For a vector $v$, we use $\| v \|$ to denote its $\ell_2$-norm.

\textbf{Threat Model.}
We consider the following use scenario of watermarks between a third-party user Alice, e.g., an artist, a generative model provider Bob, e.g., DALLE-2 from OPENAI~\cite{ramesh2022hierarchical}, and an adversary Cathy.

\begin{itemize}
    \item Alice selects a diffusion model (DM)  from Bob's API interface and sends an input  (e.g., a prompt for text-to-image diffusion models) to Bob for generating images;
    \item Bob generates images $X \in \mathcal{X}$ based on Alice's input and return $X$ to Alice;
    \item Alice embeds a watermark into the originally generated content $X$, denoted as $X^{\prime} \in \mathcal{X} $ and release to the public;
    \item Adversary Cathy applies (adversarial) image transformation(s), e.g., rotating and cropping, on  $X^{\prime}$ to obtain a modified version $\widetilde{X}^{\prime} (\in \mathcal{X} )$;
    \item Alice decides if $\widetilde{X}^{\prime} \in \mathcal{X}$ was generated by herself or not.
   \end{itemize} 

\textbf{Problem Formulation.} From the above, the watermark problem for Alice essentially boils down to a binary classification or hypothesis testing problem: $$
\begin{array}{l}
\texttt{H}_0: \widetilde{X}^{\prime} \text{ was generated by Alice (Watermarked) }; \\
\texttt{H}_1: \widetilde{X}^{\prime} \text{ was NOT generated by Alice (Unwatermarked) }.
\end{array}
$$
To address this problem, Alice will build a detector given by 
\begin{equation}\label{eq:decision_function}
g(X; \mathcal{V}_\theta) =
\begin{cases}
    1 (\text{Watermarked}) & \text{if }\mathcal{V}_{\theta}(X) \geq \tau, \\
    0 (\text{Unwatermarked}) & \text{if }\mathcal{V}_{\theta}(X) < \tau,
\end{cases}
\end{equation}

where $\tau \in \mathbb{R}$ is a threshold value and $\mathcal{V}_\theta$ (to be defined later) is a scoring function which takes the image input and output a value in $[0, 1]$ to indicate its chance of being a sample generated by Alice.

\begin{remark}[Watermarks can be generated by Alice and/or Bob.]
In the above, we describe a threat model based on Alice's viewpoint. However, we emphasize that this does not prevent Bob, the model provider, from adding watermarks after the generation process. In fact, Bob can employ similar procedures as outlined earlier to insert watermarks, as our framework is applicable to third-party users, including Bob, despite our narrative emphasis on Alice's perspective.
\end{remark}

\textbf{Alice's Goals.}
Alice's objective is to design watermarking algorithms that fulfill the following objectives:
\textbf{(1) Quality:}
 the quality of watermarked images should closely match that of the original, unwatermarked images;
\textbf{(2) Identifiability:}
both watermarked and unwatermarked content should be accurately distinguishable;
\textbf{(3) Robustness:}
the watermark should be resilient against various image manipulations.

\textbf{Cathy's (Adversary) Goals.}
Cathy aims to design attack algorithms to meet the following objectives:
\textbf{(1) Watermark Removal:} the watermarks embedded by Alice can be effectively eliminated after the attacks;
\textbf{(2) Quality:} the attacks cannot significantly alter the images.


\section{RAW}

In this section, we formally introduce our RAW framework. At a high level, the RAW framework comprises two consecutive stages: a training stage and an inference stage. In the following subsection, we first provide an in-depth description of the training stage.

\subsection{Training stage}
Suppose Alice obtains a batch of diffusion model-generated images. The unwatermarked data are denoted as $\D^{\textrm{uwm}} \triangleq \left\{X_i\right\}_{i=1}^n$ for $i= 1, \ldots,n$.
Alice will need to embed watermarks into these images to protect intellectual property.

\begin{definition}[Watermarking Module]
A watermarking module is a mapping $\mathcal{E}_{\boldsymbol{w}}(\cdot): \mathcal{X} \mapsto \mathcal{X}$ parametereized by $\boldsymbol{w} \in \mathbb{R}^{d_1}.$
\end{definition}

The watermarking module can take the form of an encoder with an attention mechanism, as seen in the RivaGan~\cite{zhang2019robust}, or it can involve Fast Fourier Transformation (FFT) followed by frequency adjustments and an inverse FFT, as employed in DwtDct. 

In our RAW framework, we propose to add two distinct watermarks into both frequency and spatial domains:
\begin{equation}\label{watermarking}
    \mathcal{E}_{\boldsymbol{w}}(X) = \mathcal{F}^{-1}(\mathcal{F}(X) + c_1 \times {\color{red} v}) + c_2 \times {\color{blue} u},
\end{equation}
where ${\color{red} v}, {\color{blue} u} \in \mathcal{X}$ are two watermarks, $c_1, c_2 >0$ determine the visibility of these watermarks, and $\mathcal{F}$($\mathcal{F}^{-1}$)  represents the  Fast Fourier Transformation (FFT) (inverse FFT), respectively. For simplicity of notation, in the rest of this paper, we will denote $\boldsymbol{w} \triangleq [u, v] $.

The rationale for adopting the above approach is to simultaneously enjoy the distinct advantages offered by watermarks in both domains. In particular, the incorporation of watermark patterns in the frequency domain has been widely recognized for its effectiveness against certain image manipulations such as translations and resizing. Moreover, our empirical validation corroborates the improved robustness of spatial domain watermarking in the presence of noise perturbations. A more detailed discussion is provided in Section~\ref{sec: further_dis}.

We denote the watermarked dataset
$\mathcal{E}_{\boldsymbol{w}}$ to be $\D^{\textrm{wm}} \triangleq \left\{\mathcal{E}_{\boldsymbol{w}}(X_i)\right\}_{i=1}^n$ for $i = 1,\ldots, n.$
Alice now wishes to distinguish the combined dataset  $\D \triangleq \D^{\textrm{uwm}} \bigcup \D^{\textrm{wm}}$ with a verification module, which is a binary classifier.

\begin{definition}[Verification Module]
A verification module is a mapping $\mathcal{V}_\theta(\cdot): \mathcal{X} \mapsto    [0,1]$ parameterized by $\theta \in \mathbb{R}^{d_2}.$
\end{definition}

The score generated by the verification module for an input image can be understood as 
the chance of this image being generated by Alice.

To fulfill Alice's first two goals, Alice will consider \textit{jointly} training the watermarking and verification modules parameterized by $\boldsymbol{w}$ and $\theta$, respectively, with the following loss function:
\begin{equation}\label{eq: simple-bce}
    \operatorname{BCE}(\D) \triangleq \sum_{X \in \D} Y\log(\mathcal{V}_{\theta}(X)) + (1-Y)\log(1-\mathcal{V}_{\theta}(X)),
\end{equation}
where $X$ is the training image and $Y \in \{0,1\}$ is the label indicting $X$ is watermarked or not.

Recall that Alice also aims to enhance the robustness of the watermark algorithms. As a result, we consider transforming the combined datasets with different data augmentations $\mathcal{M}_1, \ldots, \mathcal{M}_m$ to obtain $\D^1 \triangleq \mathcal{M}_1(\D), \ldots, \D^m \triangleq \mathcal{M}_m(\D)$, respectively. Here, the data augmentations $\mathcal{M}_1, \ldots, \mathcal{M}_m$ are defined as follow.

\begin{definition}[Modification Module]
An image modification module is a map $\mathcal{M}: \mathcal{X} \mapsto \mathcal{X}.$
\end{definition}


To sum up, the overall loss function for our RAW framework is specified as:
\begin{equation}\label{overal_loss}
\mathcal{L}_{\textrm{raw}} \triangleq \underbrace{\operatorname{BCE}(\D)}_{\mathcal{L}_0} + \underbrace{\sum_{k=1}^m\operatorname{BCE}(\mathcal{D}^k) }_{\mathcal{L}_\text{Aug}},
\end{equation}
where $\operatorname{BCE}(\cdot)$ denotes the binary cross entropy loss as specified in Equation~(\ref{eq: simple-bce}). The loss function above is composed of two terms: ${\mathcal{L}_0}$, which corresponds to the cross-entropy on the original combined datasets $\D$, and ${\mathcal{L}_\text{Aug}}$, signifying the cross-entropy on the augmented datasets $\D^1, \ldots, \D^m$. In our experiments, inspired by contrastive learning such as those presented in~\cite{chen2020simple, khosla2020supervised}, we adopt a two-view data augmentation approach by setting $m=2$.

\subsubsection{Overall Training Algorithm}

We describe the overall training algorithm below, with pseudocode summarized in Algorithm~\ref{algo: FF+ Digital}.
We consider conducting the following two steps alternatively.
\begin{itemize}
    \item Optimize the verification module $\mathcal{V}_{\theta}$ based on the overall loss $\mathcal{L}_{\textrm{raw}}$ by stochastic gradient descent (SGD):
    \begin{equation}
        \theta_{t+1} \leftarrow  \theta_t - \mu_t \nabla_{\theta}\mathcal{L}_{\textrm{raw}}(\theta_t, \boldsymbol{w}),
    \end{equation}
    where $\mu_t>0 $ is the step size at each step $t$.
   \item Optimize the watermark $\boldsymbol{w}$ based on $\mathcal{L}_0$ with sign-based stochastic gradient descent (SignSGD):
   \begin{equation}\label{watermark_update}
         \boldsymbol{w}_{t+1} \leftarrow \boldsymbol{w}_t- \nu_t \operatorname{sign}\left(\nabla_{\boldsymbol{w}} \mathcal{L}_{\textrm{raw}}(\theta, \boldsymbol{w}_t)\right),
   \end{equation}
   where $\operatorname{sign}(\cdot)$ is the signum function that outputs the sign of each of its components, and $\nu_t >0 $ is the step size.
\end{itemize}

In the watermark update, Equation~(\ref{watermark_update}), we opt for signSGD over vanilla SGD. This choice is motivated by several existing empirical observations that (sign-based) first-order methods could yield improved training and test performance in the context of data-level optimization problems in deep learning~\cite{madry2017towards,liu2021probabilistic}. Consequently, we adhere to this convention and utilize SignSGD for watermark optimization.

\begin{algorithm}[!htb]
		\centering
		\caption{Training Algorithms for RAW}\label{algo: FF+ Digital}
		\footnotesize
		\begin{algorithmic}[1]
			\renewcommand{\algorithmicrequire}{\textbf{Input:}}
			\renewcommand{\algorithmicensure}{\textbf{Output:}}
			\REQUIRE 	
			(I) Image sets generated from a diffusion model $\{X_i\}_{i=1}^n$; (II) watermark visibility parameter $c_1, c_2$; (III) learning rates $\{\mu_t\}_{t=1}^T, \{\nu_t\}_{t=1}^T.$
            \\
            \textit{Initialize}: (1) a verification module $\mathcal{V}_{\theta}: \mathcal{X} \mapsto [0,1]$, (2) a watermarking module: $\mathcal{E}_{\boldsymbol{w}}(X) = \mathcal{F}^{-1}(\mathcal{F}(X) + c_1 \times v) + c_2 \times w $ with each entries in $u,v \in \mathcal{X}$ initialized as IID uniform random variables.
			 \\
			\hrulefill
           \FOR{$i=1$ to $T$}
            \STATE Clipping the watermarked data to be within the range $[0,1]$;
            \STATE Given $\mathcal{V}_{\theta}$, optimizing $\boldsymbol{w} $ based on $\mathcal{L}_{\text{raw}}$ with SignSGD;
            \STATE Given the watermark $\boldsymbol{w} $, updating $\theta$ based on $\mathcal{L}_{\text{raw}}$ with SGD;
            \ENDFOR
            \\\hrulefill\\
			
			\ENSURE  (1) The verification module $\mathcal{V}_{\theta}$; (2) Watermarking method $\mathcal{E}_{\boldsymbol{w}}$

		\end{algorithmic}
		
	\end{algorithm}

\vspace{-0.3cm}
\subsubsection{Further Discussions}\label{sec: further_dis}
We now elaborate on two pivotal aspects of our watermark designs and overarching training algorithms: \textbf{(I)} the joint training scheme for watermarking and verification modules, and \textbf{(II)} the integration of spatial-domain watermarks.

\begin{figure}[!htb]
  \centering
  \begin{subfigure}{0.23\textwidth}
    \centering
\includegraphics[width=\textwidth]{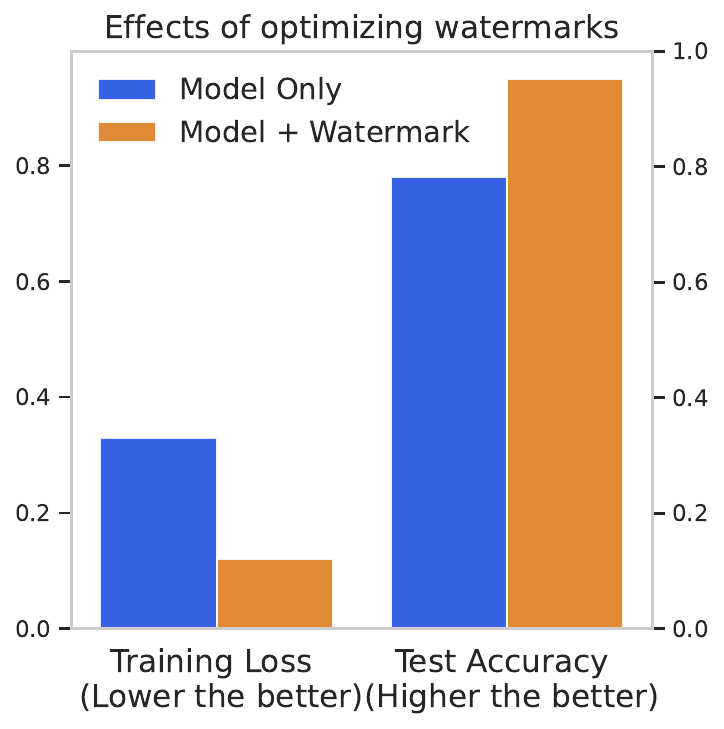}
    \caption{}
    \label{fig: wm_train}
  \end{subfigure}
  \hfill
  \begin{subfigure}{0.23\textwidth}
    \centering
\includegraphics[width=\textwidth]{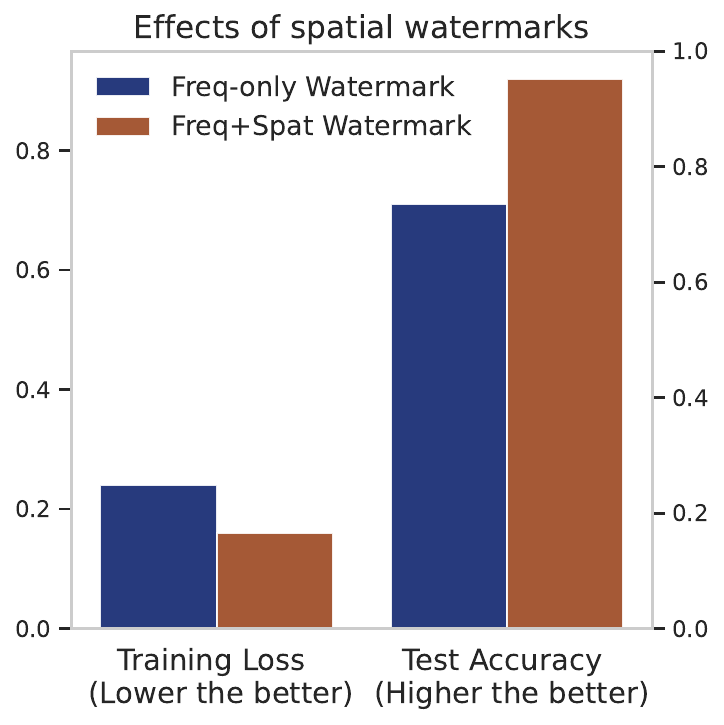}
    \caption{}
    \label{fig: wm_spa}
  \end{subfigure}
  \vspace{-0.3cm}
  \caption{Effects of (a) jointly training watermarks and models and (b) using spatial watermarks on training loss and test accuracy.}
  \label{fig:overall}
\end{figure}

\textbf{(I) The joint training scheme for watermarking and verification modules.}
Theoretically, using standard arguments from classical learning theory~\cite{vapnik1999overview}, it can be shown that training both the watermarking and the verification modules to distinguish between watermarked and unwatermarked data will not lead to a test accuracy worse than when the watermark is fixed, and only the model is trained.
From a practical perspective, the initially randomly initialized watermarks may not align well with specific training data, emphasizing the need to optimize watermarks for distinct data scenarios. Our empirical observations support this notion, as evidenced in Figure~\ref{fig: wm_train}, where the joint training scheme leads to a significantly higher test accuracy and lower training loss compared with the scenario where the watermark is fixed.

\textbf{(II) The inclusion of spatial domains.}
Classical methods for embedding watermarks primarily introduce them into the frequency domains of images~\cite{cox2007digital}. However, it has been empirically observed that such watermarks are susceptible to manipulations, such as Gaussian noise~\cite{wen2023tree}.
To overcome this vulnerability, we draw inspiration from the model reprogramming literature~\cite{chen2022model}, where watermarks are incorporated into the spatial domain to enhance accuracy in distinguishing in- and out-distribution data~\cite{wang2022watermarking}. Consequently, we explore the integration of watermarks into the spatial domain (in addition to the frequency domain), as outlined in Equation~(\ref{watermarking}). We empirically observed that including spatial watermarks could significantly boost the test accuracy of the trained verification module under Gaussian-noise manipulations on test data, as depicted in Figure~\ref{fig: wm_spa}.

\subsection{Inference Stage}
In this section, we present a generic approach for Alice to obtain provable guarantees on the False Positive Rate (FPR) when using the previously trained $\mathcal{V}_\theta$ on test images, even amidst minor perturbations. 

To begin with, we first examine a scenario where the future test data $X_{\text{test}} \in \mathcal{X}$ adheres to an IID pattern with the watermarked data $\D^{\textrm{wm}}$ generated by Alice, without undergoing any image modifications. In this case, Alice can employ conformal prediction to establish provable guarantees on the FPR. The main idea is that, by utilizing the trained $\mathcal{V}_\theta$ as a scoring mechanism, the empirical quantile of the watermarked data's distribution will converge to the population counterpart. This convergence is guaranteed by the uniform convergence of cumulative distribution functions (CDFs).
To be more specific, by setting the threshold $\tau$, defined in Equation~(\ref{eq:decision_function}), to be the $\alpha$-quantile (with finite-sample corrections) of predicted scores for watermarked data $\mathcal{V}_\theta(\D^{\textrm{wm}}) \triangleq \{\mathcal{V}_\theta(\mathcal{E}_{\boldsymbol{w}}(X_i))\}_{i=1}^n$, it can be shown~\cite{lei2014distribution} that the probability of the resulting detector $g$ misclassifying a watermarked image $X_{\textrm{test}}$ is upper bounded by $\alpha$, with high probability, under the condition that $X_{\textrm{test}}$ is IID with $\D^{\textrm{wm}}.$ For the sake of completeness, a rigorous statement and its proof are provided in the Appendix.

The above argument assumes that the future test image $X_\text{test}$ follows an IID pattern with the original watermarked data $\D^{\textrm{wm}}$. However, if the test image $X_\text{test}$ undergoes manipulation or attack, denoted by $\mathcal{A}(X_{\textrm{test}})$, with  $\mathcal{A}: \mathcal{X} \mapsto \mathcal{X}$ being an adversarial image manipulation, then it can deviate from the distribution of $\D^{\textrm{wm}}$. This deviation from IID will render the previous argument invalid.
Moreover, in practice, Alice is unaware of the adversarial transformation $\mathcal{A}$ employed by the attacker. Consequently, Alice has limited information about future test data, making it even more challenging to control the FPR.

To address this problem, we propose to consider a robust version of the originally trained $\mathcal{V}_\theta$, denoted as $\mathcal{V}_{\tilde{\theta}}$, such that $X$ and $\mathcal{A}(X)$ stay close under $\mathcal{V}_{\tilde{\theta}}$, namely
\begin{equation}\label{dis}
|\mathcal{V}_{\tilde{\theta}}(X) - \mathcal{V}_{\tilde{\theta}}(\mathcal{A}(X))| \leq \eta,
\end{equation}
for all $X$ and a small $\eta >0$.
The reason for finding such $\mathcal{V}_{\tilde{\theta}}$  is because we can  relate $\mathcal{V}_{\tilde{\theta}}(\mathcal{A}(X_{\text{test}}))$ back to $\mathcal{V}_{\tilde{\theta}}(X_{\text{test}})$ which is IID with $\mathcal{V}_{\tilde{\theta}}(\D^{\textrm{wm}})$ (accessible to Alice)  to establish the FPR with previous arguments.

How can we develop the robust version from the base $\mathcal{V}_\theta$? The following result from the randomized smoothing technique offers a possible solution. Denote $\mathcal{N}(\mu, \Sigma)$ to be the normal distribution with mean $\mu$ and covariance $\Sigma$ respectively, and
$\Phi^{-1}(\cdot)$ to be the inverse of the cumulative distribution function of a standard normal distribution.
\begin{lemma}[\cite{salman2019provably}]
Let $h: \mathbb{R} \rightarrow[0,1]$ be a continuous function. Let $\sigma>0$,  and $H(x) = \underset{Z \sim \mathcal{N}\left(0, \sigma^{2} I\right)}{\mathbb{E}}[h(X+Z)]$. Then the function $\Phi^{-1}(H(X))$ is $\sigma^{-1}$-Lipschitz. 
\end{lemma}

The above result suggests that for \textit{any} base verification module (classifier)  $\mathcal{V}_{\theta}$, we can obtain a smoothed version with 
\begin{equation}\label{smoothing}
    \mathcal{V}_{\tilde{\theta}}(X) =  \Phi^{-1}\biggl(\underset{Z \sim \mathcal{N}\left(0, \sigma^{2} I\right)}{\mathbb{E}}[\mathcal{V}_{\theta}(X+Z)]\biggr),
\end{equation}
and it is guaranteed that 
$
 |\mathcal{V}_{\tilde{\theta}}(X) - \mathcal{V}_{\tilde{\theta}}(Y)| \leq \sigma^{-1}\|X-Y\|,$
for any $X,Y \in \mathcal{X}.$
Suppose the attacker employs an adversarial attack $\mathcal{A}$ such that $
\|X - \mathcal{A}(X)\| \leq \gamma$. We have
\begin{equation}
|\mathcal{V}_{\tilde{\theta}}(X) - \mathcal{V}_{\tilde{\theta}}(\mathcal{A}(X))| \leq \frac{\gamma}{\sigma}.
\end{equation}

\begin{remark}[$\mathcal{A}$ can not be excessively adversarial]
We emphasize that the transformation $\mathcal{A}$ should not be excessively adversarial. In other words, the parameter $\gamma$ should be a very low value for both theoretical and practical reasons.
From a theoretical perspective, an overly adversarial transformation $\mathcal{A}$ can result in trivial TPR/FPR. For instance, if watermarked images are transformed into a completely uniform all-white or all-black state, it becomes impossible to detect the watermark.
From a practical standpoint, an excessively adversarial transformation $\mathcal{A}$ tends to overwrite the original content within the images. This directly contradicts the intentions of attackers and may not achieve the desired stealthy modifications.
\end{remark}

\subsubsection{Overall Inference Algorithm}
Given a pair of $(\mathcal{E}_{\boldsymbol{w}}$, $\mathcal{V}_{\theta}$), a desired robust range $\gamma>0$, and a smoothing parameter $\sigma >0$, Alice now will set the thresholding value $\tau$, as introduced in Equation~(\ref{eq:decision_function}),  to satisfy:
\begin{equation}\label{quantile_selection}
    \hat{F}\bigl(\tau - \frac{\gamma}{\sigma}\bigr) =  \alpha - \sqrt{(\log(2/\delta)/(2n))},
\end{equation}
 where $\delta \in (0,1)$ is a violation rate describing the probability that the $\operatorname{FPR}$ exceeds $\alpha$, and $\hat{F}$ is the empirical cumulative distribution function of $\{\mathcal{V}_{\tilde{\theta}}(\mathcal{E}_{\boldsymbol{w}}(X_i))\}_{i=1}^n$, where 
 $$\mathcal{V}_{\tilde{\theta}}(\mathcal{E}_{\boldsymbol{w}}(X_i)) \de 
 \Phi^{-1}\biggl(\underset{Z \sim \mathcal{N}\left(0, \sigma^{2} I\right)}{\mathbb{E}}[\mathcal{V}_{\theta}(\mathcal{E}_{\boldsymbol{w}}(X_i)+Z)]\biggr) . $$

The next result shows that if a future test input comes from the same distribution as the watermarked data $\D^{\textrm{wm}}$, the above procedure can be configured to achieve any pre-specified false positive rate $\alpha$ with high probability. 

\begin{theorem}[Certified FPR of $g$ based on threshold in Equation~(\ref{quantile_selection})]\label{fpr}
Given any watermarked dataset $\D^{\text{wm}}$ and its associated verification module $\mathcal{V}_\theta$, suppose that the test data $(X_{\textrm{test}}, Y_{\textrm{test}})$ are IID drawn from the distribution of $\D^{\text{wm}}$.
Given any $\delta \in (0,1)$ and $\gamma >0$, for any (adversarial) image transformations $\mathcal{A}$ such that $\|\mathcal{A}(X) - X\| \leq \gamma$ for all $X \in \mathcal{X}$,  the detector $g(\cdot)$ introduced in Equation~(\ref{eq:decision_function}), with the threshold $\tau$ as specified in Equation~(\ref{quantile_selection}) satisfies 
 $$ \mathbb{P}\biggl( g(\mathcal{A}(X_{\textrm{test}})) = 0 \, (\text{Unwatermarked}) \mid X_{\textrm{test}} \sim \D^{\text{wm}} \biggr) \leq  \alpha,$$ 
 with probability at lease $1-\delta$ for any $\alpha \in (0,1)$ such that $\alpha >  \sqrt{(\log(2/\delta)/(2n))}.$
 \end{theorem}

The above result shows that by using the decision rule as specified in Equation~(\ref{quantile_selection}), Alice can obtain a provable guarantee on the Type I error rate in terms of detecting future test input $X_{\text{test}}$ even $X_{\text{test}}$ is adversarially perturbed within $\gamma$-range (as measured by $\ell_2$-norm), under the condition that the future test input $X_{\text{test}}$ is independently and identically distributed as the $\D^{\text{wm}}$, namely watermarked samples generated by the artist.

\section{Experiments}

In this section, we conduct a comprehensive evaluation of our proposed RAW, assessing its detection performance, robustness, watermarking speed, and the quality of watermarked images. Our findings reveal significantly enhanced robustness in RAW while preserving the quality of generated images. Furthermore, a substantial reduction in watermark injection time, up to $30 \times$ faster, indicates the suitability of RAW for on-the-fly deployment. All the experiments were conducted on cloud computing machines equipped with Nvidia
Tesla A100s.

\begin{table*}[!htb]
\caption{Summary of main results. The `Fixed Model' column indicates whether the method alters the underlying generative models. AUROC (Nor) denotes the AUROC performance without image manipulations or adversarial attacks. AUROC (Adv) represents the average performance across nine distinct image manipulations and attacks. The `Encoding Speed' column denotes the efficiency of watermark injection post-training, measured in seconds per image.}
\begin{center}
\scalebox{0.95}{
\begin{tabular}{p{2cm}ccccccccc}
\toprule
\multicolumn{1}{c}{Dataset} &
\multicolumn{1}{c}{Method} &
\multicolumn{1}{c}{Fixed Model} &
\multicolumn{1}{c}{AUROC (Nor) $\uparrow$} &
\multicolumn{1}{c}{AUROC (Adv) $\uparrow$} &
\multicolumn{1}{c}{FID $\downarrow$} &
\multicolumn{1}{c}{CLIP $\uparrow$} & \multicolumn{1}{c}{Encoding Speed $\downarrow$}
\\
\cmidrule(lr){1-1} \cmidrule(lr){2-2} \cmidrule(lr){3-3} \cmidrule(lr){4-4} \cmidrule(lr){5-5} \cmidrule(lr){6-6} \cmidrule(lr){7-7} \cmidrule(lr){8-8} 
 & DwTDcT & \textcolor{blue}{\ding{51}} & ${0.83}$ & $0.54$ & $25.10$ & $0.359$ & $0.048$  \\
\multirow{1}{3cm}{MS-Coco} & DwTDcTSvd & \textcolor{blue}{\ding{51}} & ${0.98}$ & $0.75$ & $25.21$ &  $0.361$ & $0.122$ \\
 & RivaGan & \textcolor{blue}{\ding{51}} & ${0.99}$ & $0.81$ & $24.87$ & $0.359$  & $1.16$  \\
  & RAW (Ours) & \textcolor{blue}{\ding{51}} & ${0.98}$ & $\mathbf{0.92}$ & ${24.75}$ & $0.360$ & $\mathbf{0.0051}$ \\
  \cmidrule(lr){1-1} \cmidrule(lr){2-2} \cmidrule(lr){3-3} \cmidrule(lr){4-4} \cmidrule(lr){5-5} \cmidrule(lr){6-6} \cmidrule(lr){7-7} \cmidrule(lr){8-8} 
 & DwTDcT & \textcolor{blue}{\ding{51}} & ${0.81}$ & $0.55$ & $4.63$ & $0.427$ & $0.048$  \\
\multirow{1}{3cm}{DBdiffusion} & DwTDcTSvd & \textcolor{blue}{\ding{51}} & ${0.99}$ & $0.78$ & $4.61$ &  $0.421$  & $0.110$ \\
 & RivaGan & \textcolor{blue}{\ding{51}} & ${0.99}$ & $0.82$ & $4.82$ & $0.424$   & $1.87$ \\
  & RAW (Ours) & \textcolor{blue}{\ding{51}} & ${0.98}$ & $\mathbf{0.90}$ & ${5.17}$ & $0.425$ & $\mathbf{0.0078}$    \\
\bottomrule
\end{tabular}
}
\end{center}
\label{tpr_result}
\end{table*}

\begin{table*}[h]
\centering
\caption{AUROC performance of state-of-the-art methods under 9 (adversarial) image manipulations (Rotation $90^\circ$, Cropping and resizing 70\%,  Gaussian Blur with a kernel size of $(7,9)$ and bandwidth of $4$, Noise with IID mean Gaussian $\sigma = 0.05$, Jitter with brightness factor $0.6$, JPEG compression with quality $50$, and 3 adversarial attacks for removing watermarks) with Algo.~\ref{algo: FF+ Digital}.}
\begin{tabular}{ccccccccc}
\toprule

\multicolumn{1}{c}{Datasets} & \multicolumn{4}{c}{MS-COCO} & \multicolumn{4}{c}{DBDiffusion} \\
\cmidrule(lr){2-5} \cmidrule(lr){6-9}
\multicolumn{1}{l}{} & \multicolumn{1}{c}{DwtDct} & \multicolumn{1}{c}{DwtDctSvd} & \multicolumn{1}{c}{{RivaGan}}  & \multicolumn{1}{c}{{RAW (Ours)}} & \multicolumn{1}{c}{DwtDct} &  \multicolumn{1}{c}{DwtDctSvd} & \multicolumn{1}{c}{{RivaGan}}  & \multicolumn{1}{c}{{RAW (Ours)}} \\

\cmidrule(lr){1-1} \cmidrule(lr){2-2} \cmidrule(lr){3-3} \cmidrule(lr){4-4} \cmidrule(lr){5-5} \cmidrule(lr){6-6} \cmidrule(lr){7-7} \cmidrule(lr){8-8} \cmidrule(lr){9-9}
\multicolumn{1}{l}{JPEG $50$} & $0.612$ & $0.995$ & $0.996$ & ${0.914}$ & $0.503$ & $0.954$  & $0.997$ & $0.999$ \\
\multicolumn{1}{l}{Rotation $90^\circ$} & $0.508$ & $0.547$ & ${0.391}$ & ${0.956}$ & $0.471$ & $0.541$ &  ${0.381}$ & ${0.824}$ \\
\multicolumn{1}{l}{Cropping $70\%$} & $0.640$ & $0.521$ & ${0.990}$  & ${0.957}$ & $0.651$ & $0.613$ &   ${0.991}$ & ${0.843}$ \\
\multicolumn{1}{l}{Gaussian Blur} & $0.524$ & $0.916$  & ${0.999}$  &  ${0.936}$ & $0.533$ & $0.994$ & ${0.999}$ & ${0.999}$ \\
\multicolumn{1}{l}{Gaussian Noise} & $0.475$ & $0.763$ & ${0.999}$ & ${0.902}$ & $0.844$  & $0.988$ & ${0.999}$ & ${0.999}$\\
\multicolumn{1}{l}{Jittering} & $0.651$ & $0.782$ & ${0.987}$ &${0.956}$ & $0.467$  & $0.688$ & ${0.987}$ & ${0.999}$\\
\multicolumn{1}{l}{VAE Att1} & $0.502$ & $0.728$ & ${0.628}$ & ${0.895}$ & $0.488$  & $0.751$ & $0.673$ & ${0.801}$ \\
\multicolumn{1}{l}{VAE Att2} & $0.483$ & $0.775$ & ${0.671}$ & ${0.912}$ & $0.498$  & $0.725$ & $0.630$ & ${0.810}$\\
\multicolumn{1}{l}{Diff Att} & $0.498$ & $0.713$ & $ 0.698$& ${0.828}$ & $0.507$ & $0.733$ & $0.703$ & ${0.824}$\\
\multicolumn{1}{l}{Average} & $0.543$ & $0.748$ & $0.817$ & ${\mathbf{0.918}}$ & $0.551$ & $0.776$ & $0.818$  & ${\mathbf{0.901}}$ \\
\bottomrule
\end{tabular}
\label{adv_results}
\end{table*}

\subsection{Experimental setups}

\textbf{Datasets} 
\textbf{(1)}  In line with the previous work~\cite{wen2023tree}, we employ Stable Diffusion-v2-1~\cite{rombach2022high}, an open-source text-to-image diffusion model, with DDIM sampler, to generate images. All the prompts used for image generation are sourced from the MS-COCO dataset~\cite{lin2014microsoft}.
\textbf{(2)} 
We further evaluate our RAW utilizing DBdiffusion~\cite{wang2022diffusiondb}, a dataset consisting of 14 million images generated by Stable Diffusion. This dataset encompasses a wide array of images produced under various prompts, samplers, and user-defined hyperparameters, featuring both photorealistic and stylistic compositions, including paintings. For each dataset, 500 images are randomly selected for training, and subsequently, we evaluate the trained watermarks and associated models on 1000 new, unwatermarked images and their watermarked versions.

\subsection{Clean detection performance and image generation quality}
We assess (1) the detection performance of our RAW under no image manipulation or adversarial attacks and (2) the quality of the watermarked images in this subsection. As a primary evaluation metric for detection performance, we follow the convention of reporting the area under the curve of the receiver operating characteristic curve (AUROC)~\cite{wen2023tree,fernandez2022watermarking,fernandez2023stable}.
To assess the quality of the generated watermarked images, following~\cite{wen2023tree}, we adopt both the Frechet Inception Distance (FID)~\cite{heusel2017gans} and the CLIP score~\cite{radford2021learning}.
All metrics are averaged across 5 independent runs.

\begin{figure}
    \centering
\includegraphics[width = 0.95\linewidth]{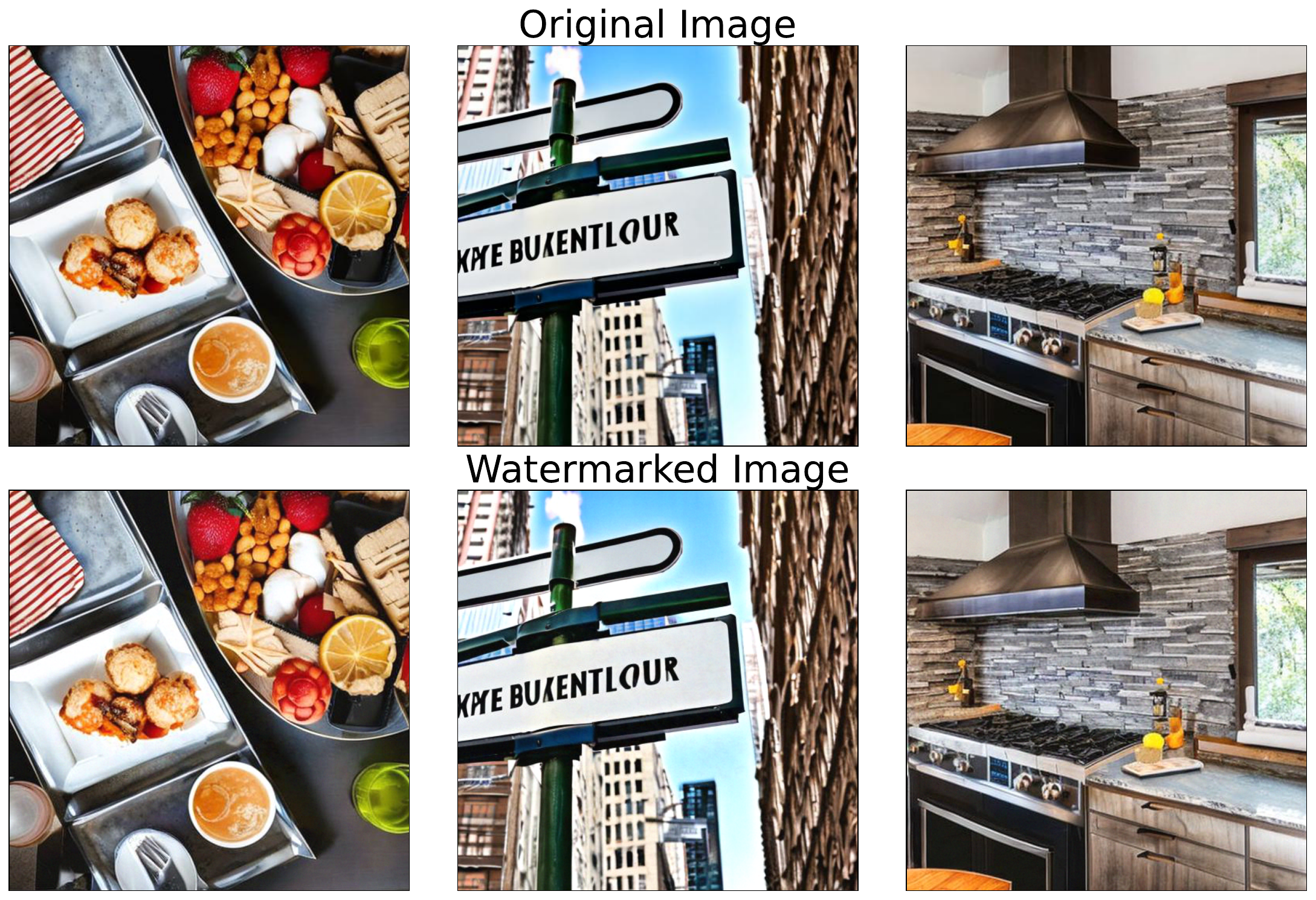}
    \caption{Examples of RAW-watermarked images (bottom row).}
    \label{fig: visual_examples}
    \vspace{-0.3cm}
\end{figure}

The results are summarized in Table~\ref{tpr_result}, and visual examples are illustrated in Figure~\ref{fig: visual_examples}. Our RAW method exhibits comparable performance to encoder-decoder-based approaches, while concurrently achieving similar FID and CLIP scores, which underscores superior image quality compared to alternative methods.

\subsection{Robust detection performance}
We assess the robustness of our proposed RAW against six common data augmentations and three adversarial attacks in this subsection. The data augmentation set comprises: color jitter with a brightness factor of $0.5$, JPEG compression with quality $50$, rotation by $90^\circ$, addition of Gaussian noise with $0$ mean and standard deviation $0.05$, Gaussian blur with a kernel size of $(7,9)$ and bandwidth $4$, and $70\%$ random cropping and resizing. These manipulations represent typical, yet rather strong, image processing operations that could potentially affect watermarks. 
Additionally, we conduct ablation studies to investigate the impact of varying intensities of these manipulations in the Appendix. 
For adversarial attacks, we evaluate our RAW against three state-of-the-art methods for removing watermarks, with two VAE-based attacks Bmshj2018~\cite{balle2018variational} (VAE Att1) and Cheng2020~\cite{cheng2020learned} (VAE Att2) from CompressAI~\cite{begaint2020compressai}, and one diffusion-model-based attacks.  All attacks were replicated by re-running publicly available codes (details in
the Appendix) with their default hyperparameters. 

The results are summarized in Table~\ref{adv_results}. Our approach demonstrates superior performance compared with alternative methods. Specifically, across both datasets, the average AUROC for our RAW increased by $70\%$ and $13\%$ for nine image manipulations/attacks, surpassing frequency- and encoder-decoder-based methods. Notably, for image manipulation involving a $90^\circ$ rotation and adversarial attacks, the AUROC of our RAW is around $0.9$, while other methods hover around $0.6$, showing a substantial performance gap that underscores the robustness and effectiveness of our approach in handling this specific manipulation scenario. 

\subsection{Watermark embedding speed}
In this section, we investigate the time costs needed to embed watermarks into images. We note that the watermark injection process occurs post-training. Therefore, our watermark injections only necessitate one FFT, two additions, and another inverse FFT.
In Table~\ref{cpu_time}, we present CPU times for watermark injection into different image quantities. Notably, our method shows significant efficiency gains, approximately 30 times faster than the Frequency-based method. This is attributed to streamlined batch operations in our RAW. This highlights the suitability of our approach for on-the-fly deployment.

\begin{table}[!htb]
\centering
 \caption{CPU time (seconds) elapsed for embedding watermarks.}
 \vspace{-0.1cm}
\begin{tabular}{*{4}{c}}

\toprule
 \multicolumn{1}{c}{} & 
 \multicolumn{1}{c}{5 images} & 
 \multicolumn{1}{c}{100 images} &
  \multicolumn{1}{c}{500 images} 
  \\
\midrule
  \multicolumn{1}{c}{DwtDct} & 
 \multicolumn{1}{c}{$0.27$} &
 \multicolumn{1}{c}{$4.8$} &
 \multicolumn{1}{c}{$24.5$} 
 \\
   \multicolumn{1}{c}{DwtDctSvd} & 
 \multicolumn{1}{c}{$0.64$} &
 \multicolumn{1}{c}{$12.2$} &
 \multicolumn{1}{c}{$60.1$} \\
    \multicolumn{1}{c}{RivaGAN} & 
 \multicolumn{1}{c}{$5.52$} &
 \multicolumn{1}{c}{$116$} &
 \multicolumn{1}{c}{$>500$} \\
    \multicolumn{1}{c}{RAW (Ours)} & 
 \multicolumn{1}{c}{$0.35$} &
 \multicolumn{1}{c}{$0.51$} &
 \multicolumn{1}{c}{$0.76$} \\
 \bottomrule
\end{tabular}\label{cpu_time}
\vspace{-0.1cm}
\end{table}

\subsection{Certified FPRs}
We assess the certified FPRs performance of our proposed RAW by varying the
FPR rate $\alpha$ pre-specified by Alice. 
We set the adversaril radius $\gamma = 0.001$ and the smoothing parameter $\sigma = 0.05$.
We summarize the results of five independent runs in Table~\ref{fpr_table} and report the mean (with standard error $< 0.002$).
The results demonstrate that the FPR of RAW consistently matches the theoretical
upper bounds (i.e., $\alpha$), supporting the result presented in Theorem~\ref{fpr}.

\begin{table}[!htb]
\centering
 \caption{Certified FPRs under different $\alpha$.}
\begin{tabular}{*{5}{c}}

\toprule
 \multicolumn{1}{c}{$\alpha$} &
 \multicolumn{1}{c}{$0.005$} & 
 \multicolumn{1}{c}{$0.01$} & 
 \multicolumn{1}{c}{$0.05$} &
 \multicolumn{1}{c}{$0.1$} 
  \\
\midrule
  \multicolumn{1}{c}{FPR} & 
 \multicolumn{1}{c}{$0.0042$} &
 \multicolumn{1}{c}{$0.0089$} &
 \multicolumn{1}{c}{$0.043$} & 
  \multicolumn{1}{c}{$0.095$}\\
 \bottomrule
\end{tabular}\label{fpr_table}
\end{table}

\section{Conclusion}
In this work, we present the RAW framework as a generic watermarking strategy crucial for safeguarding intellectual property and preventing potential misuse of AI-generated images. RAW introduces learnable watermarks directly embedded into images, with a jointly trained classifier for watermark detection. Its design renders RAW suitable for on-the-fly deployment post-training, providing provable guarantees on FPR even when test images are adversarially perturbed. Experimental results across datasets underscore its merits.

There are several interesting avenues for future research. One direction is exploring the maximum number of distinct watermarks that can be concurrently learned within a single training session. Another challenge is determining the optimal smoothing strategy to attain the largest certified radius.

The Appendix contains proofs, implementation details for experiments, and various ablation studies.

\newpage
{
    \small
    \bibliographystyle{ieeenat_fullname}
    \bibliography{main}
}

\end{document}